\documentclass{article}

\usepackage[nonatbib,final]{ap_2020}

\usepackage[utf8]{inputenc} 
\usepackage[T1]{fontenc}    
\usepackage{hyperref}       
\usepackage{url}            
\usepackage{booktabs}       
\usepackage{amsfonts}       
\usepackage{nicefrac}       
\usepackage{microtype}      
\usepackage{biblatex}
\usepackage{multirow}
\usepackage{graphicx}
\usepackage{xcolor}
\usepackage{adjustbox}
\usepackage{tabularx}
\usepackage{caption}
\usepackage{subcaption}
\usepackage{multirow}

\title{Alquist 4.0: Towards Social Intelligence Using Generative Models and Dialogue Personalization}

\author{
  Jakub Konr\'ad, Jan Pichl, Petr Marek,
  Petr Lorenc, Van Duy Ta, Ond\v{r}ej Kobza\\
  Faculty of Electrical Engineering, CTU Prague\\
  Prague, Czech Republic \\
  \texttt{\{konrajak, pichljan, marekp17, lorenpe2, tavanduy, kobzaond\}@fel.cvut.cz} \\
  \And
  Lenka Hýlová\thanks{\hspace{0.2cm}Research conducted at CTU Medialab} \\
  DTU Compute, Technical University of Denmark \\
  Lyngby, Denmark \\
  \texttt{lenhy@dtu.dk} \\
  \And
   Jan \v{S}ediv\'{y} \\
   CIIRC, CTU Prague \\
   Prague, Czech Republic \\
   \texttt{jan.sedivy@cvut.cz} \\
}

\begin{document}

\maketitle

\begin{abstract}
The open domain-dialogue system Alquist has a goal to conduct a coherent and engaging conversation that can be considered as one of the benchmarks of social intelligence. The fourth version of the system, developed within the Alexa Prize Socialbot Grand Challenge 4, brings two main innovations. The first addresses coherence, and the second addresses the engagingness of the conversation.
  
For innovations regarding coherence, we propose a novel hybrid approach combining hand-designed responses and a generative model. The proposed approach utilizes hand-designed dialogues, out-of-domain detection, and a neural response generator. Hand-designed dialogues walk the user through high-quality conversational flows. The out-of-domain detection recognizes that the user diverges from the predefined flow and prevents the system from producing a scripted response that might not make sense for unexpected user input. Finally, the neural response generator generates a response based on the context of the dialogue that correctly reacts to the unexpected user input and returns the dialogue to the boundaries of hand-designed dialogues.
  

The innovations for engagement that we propose are mostly inspired by the famous exploration-exploitation dilemma. To conduct an engaging conversation with the dialogue partners, one has to learn their preferences and interests---exploration. Moreover, to engage the partner, we have to utilize the knowledge we have already learned---exploitation.
  
In this work, we present the principles and inner workings of individual components of the open-domain dialogue system Alquist developed within the Alexa Prize Socialbot Grand Challenge 4 and the experiments we have conducted to evaluate them.
\end{abstract}

\section{Introduction}
The goal of the Alexa Prize competition is to advance the research of conversational AI, mainly in the human-evaluated metrics of coherence and engagement. We can roughly describe coherence as the ability of the socialbot to correctly understand and advance the dialogue, and engagement as the ability to entertain the other side of the dialogue \cite{gabriel2020further}. Both capabilities are critical for social intelligence. For the Alexa Prize Grand Challenge 4, we propose two main innovations that improve the socialbot's skills in coherence and engagement.  

The innovation for coherence is based on the novel combination of hand-designed dialogues and generative models. The hand-designed dialogues are represented as graphs consisting of nodes representing user inputs and bot responses. Nodes are organised into branching flows (\autoref{fig:stucture1}). The advantage of hand-designed dialogues is that the dialogue designers have a complete control over the flow of the dialogue. This level of control enables them to design high-quality conversations. But on the other hand, the dialogue graphs are rather rigid because they cover only the most common user inputs. Thus, when the user says something that was not anticipated, what we call out-of-domain input, the dialogue reacts with one of the responses that were prepared for the most common inputs. Such a response does not make sense in most cases.

The solution to the problem of out-of-domain inputs is made of two steps. The first step is to recognize that the dialogue is not prepared to handle the input. The second step is to produce a new response from the context of the dialogue that is coherent. We apply out-of-domain recognition to the former and the Neural Response Generator to the latter. Out-of-domain (OOD) recognition is a part of the intent classifier that recognizes that a given input is unexpected. The 
Neural Response Generator is a neural generation model trained on large dialogue corpora that produces a response based on the context of the dialogue. Combination of dialogues \cite{alquist2}, OOD recognition (\autoref{section:Intent}), and the Neural Response Generator (\autoref{section: Neural Response Generator}) allows us to utilize high-quality hand-designed dialogues while adding the necessary resilience to unexpected user inputs that is a crucial component for coherence in open-domain dialogue systems.

The proposed innovation for engagement is based on the fact that in order to entertain the conversational partner, one has to learn what entertains the partner first and then utilize the knowledge in the following conversation. This might remind us of a famous problem of computer science, the problem of exploration and exploitation.

In Alexa Prize, we are facing such a situation. It is safe to say that the socialbot is in the role of an entertainer that has zero prior knowledge about the user. Because the socialbot has zero knowledge, it has to explore the user's preferences first. However, it can't stay in a pure exploration mode for the rest of the conversation. Gradually, it has to proceed into an exploitation phase after some time, to maximize its engagement score. This philosophy is reflected in the design of the components that the fourth version of Alquist is made of.

For the exploration part, in which Alquist learns the preferences of the user, the main research and development emphasis was put on Skimmer (\autoref{section:Skimmer}; a component that extracts information the user mentions without the bot explicitly asking for it), User Profile (\autoref{section:User Profile}), and Entity and Knowledge Utilization (\autoref{section:Entity}). The mentioned components collect and organize the pieces of information mentioned by the user that are utilized in the following dialogue.

For the exploitation part, in which Alquist utilizes the knowledge about the user, the main emphasis was put on the research and development of the Dialogue Manager (\autoref{section:Dialogue Management}), Trivia Selection (\autoref{section:Trivia Selection}), Intent and Out-of-Domain classification (\autoref{section:Intent}), and the Neural Response Generator (\autoref{section: Neural Response Generator}). Those components are responsible for selecting the next action in the dialogue and response production that utilizes the knowledge about the user.

\begin{figure}
     \centering
     \hspace{1cm}
     \begin{subfigure}[b]{0.35\textwidth}
         \centering
         \includegraphics[width=\textwidth]{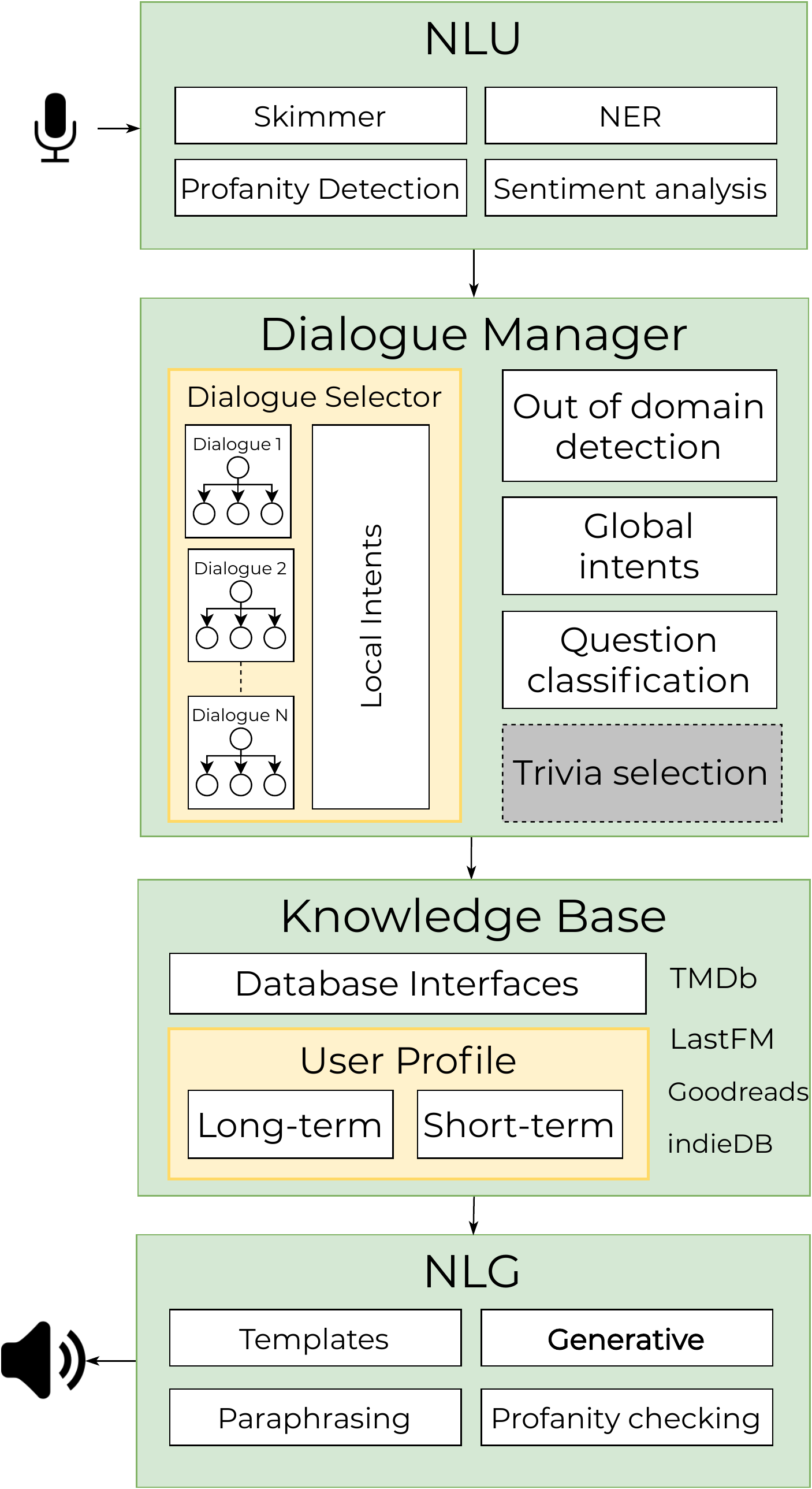}
         \caption{Exploration flow}
     \end{subfigure}
     \hfill
     \begin{subfigure}[b]{0.35\textwidth}
         \centering
         \includegraphics[width=\textwidth]{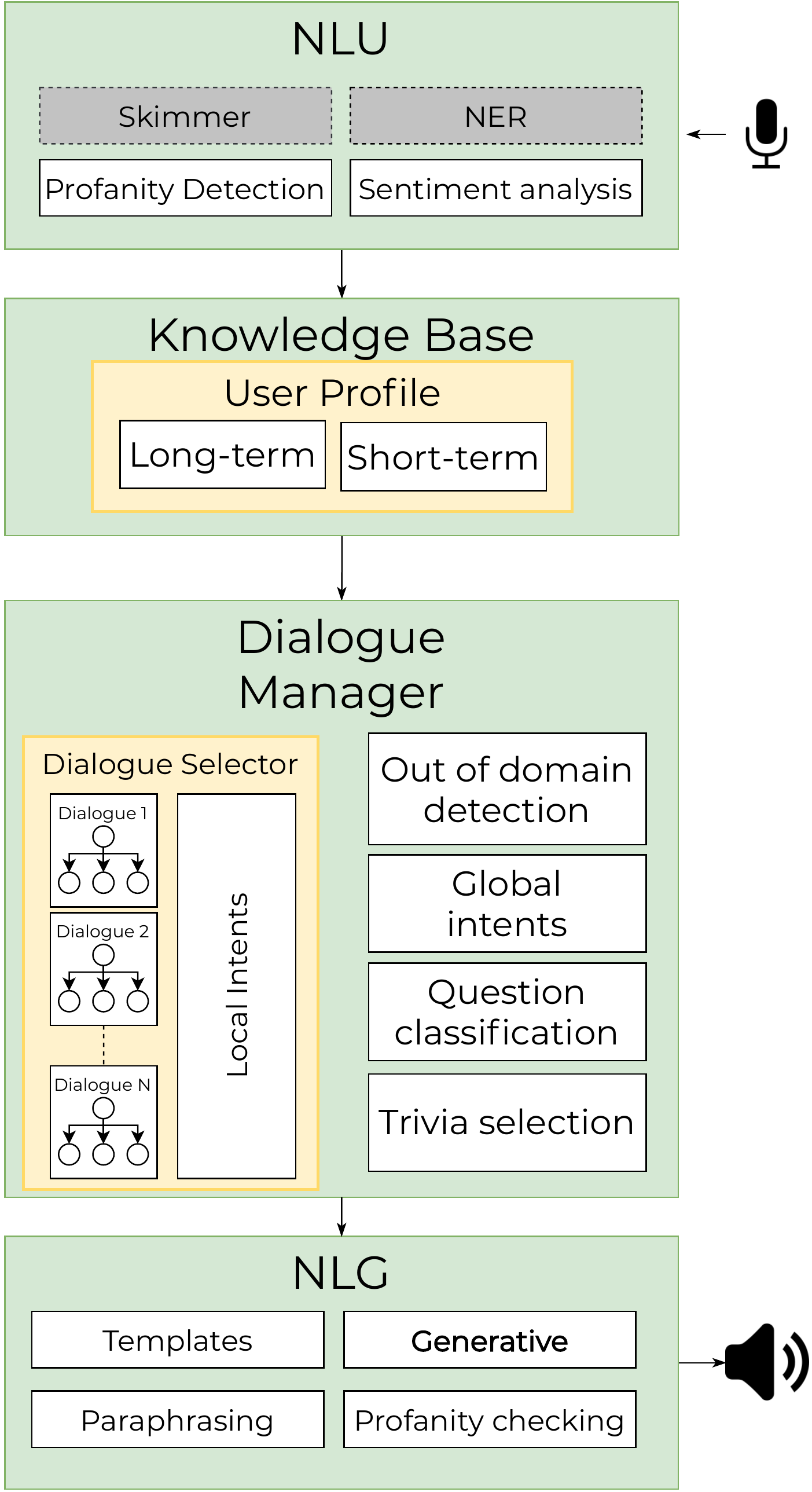}
         \caption{Exploitation flow}
     \end{subfigure}
     \hspace{1cm}
     \caption{The system components are shown in two different orders based on the dialogue strategy---exploration and exploitation. In the exploration strategy, our bot uses dialogues to extract information about the user. In the exploitation strategy, our bot utilizes the information stored in the User Profile and selects dialogues accordingly. The grey components are not essential for the corresponding strategy, but they are not disabled completely.}
     \label{fig:systemOverview}
\end{figure}

\autoref{fig:systemOverview} presents the organization of all of Alquist's components. We took \cite{alquist1,alquist2,alquist3} as our starting point. The unchanged components are described in those previous works and the newly proposed components are described in the following sections in detail. The sections are sorted by the order in which the individual components process the user input. First, the Skimmer analyses the user input for the mentioned pieces of information. The pieces of information are stored in the User Profile. Based on the values stored in the user profile, the Dialogue Management selects the next dialogue to start, or selects and presents some trivia related to the actual topic of a conversation. The dialogue is directed according to the Intent classification of the user input. And finally, if the Out-of-domain classification recognizes an unexpected user input, the Neural Response Generator produces a coherent response based on the context of the conversation.

\section{Skimmer} \label{section:Skimmer}
One of the critical aspects of each engaging conversation is working with the information mentioned by your communication partner. Such information can be expressed in two basic scenarios. The first scenario is when the bot asks a user a direct question (e.g. \textit{``Do you have a brother?''}) and stores the answer to the question. In this scenario, the bot is aware of the dialogue context, it knows what type of answer is expected, and it can store the response in the User Profile accordingly. Using the stored value, the bot can carry out a highly personalized conversation and ask relevant questions such as \textit{``How is your brother today?''}. This strategy leads to a more personalized conversation.

Since we do not want to disrupt the fluency of the dialogue by asking too many personal questions to gather information about a user, we want the bot to have the ability to extract the information from each user utterance. For example, the information about the user having a brother can be mentioned ``by the way'' in the conversation (e.g., in a movie related conversation, the user may mention \textit{``I was with my brother at the cinema yesterday.''}). We want to extract the information from the sentence regardless of the topic being discussed. For this purpose, we implemented the component called \textbf{Skimmer}. It skims through each utterance and saves the values in User Profile based on a list of rules. Each of the rules contains the following attributes:
\begin{itemize}
    \item \textit{Regular expression} - a set of patterns which must or must not (negative patterns) be contained in the utterance.
    \item \textit{User Profile attribute} - the name of the attribute where the value will be stored.
    \item \textit{Value} - the value stored in the attribute, typically \textit{true}, \textit{false}, or a matched group of the regular expression.
\end{itemize}

The component processes the user utterance in the following way. It takes each rule from the list and tries to match it with the corresponding regular expression. If it is matched, the value is stored in the specific attribute of the User Profile.

\section{User Profile} \label{section:User Profile}
An important aspect of social intelligence is the ability to retain and utilize relevant information about the conversational partner. User Profile is a unified storage of information which is relevant for the conversation. It consists of two main parts---\textbf{Long-term} User Profile and \textbf{Short-term} User Profile, also called \textit{Session-scoped} Profile. The Long-term Profile holds the information about a user across sessions. The values stored in the User Profile are used in the Dialogue Selector and additionally in individual dialogues across various topics. 

\subsection{Profile Structure}
The Long-term User Profile is divided into several sections, mostly corresponding to the topics such as movies, sports, books, etc. The additional sections contain general information about the user not specific to a topic (i.e., \textit{name}, \textit{mood}). Each section contains several attributes filled by the corresponding dialogues or by the Skimmer component on the global level. All attributes have a default value representing the state in which the bot does not know the information about the user. When the dialogue reads the default value and needs to work with it further, it asks the user a question to fill the proper attribute value.

The Short-term User Profile stores the entities discussed in the current session. It is mainly used to get the last entity discussed in the conversation or the last entity corresponding to the specific topic.

\subsection{Profile Resetting}
From the technical perspective, each user is identified by the user ID assigned to the specific Echo device. However, multiple users may use the same device and interact with the bot. At the beginning of each conversation, the bot asks for the user's name, and the following situations may happen:
\begin{itemize}
    \item The user ID has a first session with the bot --- Bot asks for the user's name and creates a new Long-term Profile.
    \item The user ID has had a session before, but the user's name is not saved --- Bot asks whether they have talked to each other before. If the user says ``yes'', the old profile is restored. Otherwise, a new one is created. In both cases, the bot asks for the user's name additionally.
    \item The user ID has had a session before, and the user's name has a value stored --- Bot asks the user to confirm his name. If it is the same user as the last session, the previous profile is used. Otherwise, a new one is created.
\end{itemize}

The Short-term User profile is reset at the beginning of each session.

\section{Entity and Knowledge Utilization}
\label{section:Entity}

The experience from the previous years of the Alexa Prize competition has shown that to have a meaningful conversation, the socialbot has to be able to chat about specific entities that naturally emerge from the conversation \cite{alquist3}. Furthermore, the conversation should both include factual information about the entity and be interested and engaged in the user's feelings and personal experiences about the topic \cite{paranjape2020neural}. To strengthen those capabilities of our socialbot, we developed a system that builds on top of classical entity detection by linking them to domain-specific publicly accessible databases rather than one general-purpose knowledge base. Then we utilize the information from the mentioned databases to continue the conversation.

\subsection{Entity Recognition}
We approach entity recognition as a sequence tagging task as described in \cite{alquist2}. We utilize the BI-LSTM \cite{huang2015bidirectional} model. We train the model on a hand-annotated dataset from the data gathered from the user utterances in the previous years of the competition. Each token in the utterance is classified into one of the defined classes, either B-\textit{type} as the beginning of an entity of a given entity type, I-\textit{type} as an inside token of a given entity type, or O as an outside token. We predict 16 entity types, such as \textit{Movie, Sport, Job, Language, Music genre,} et cetera. The predicted entity and type are then used by the dialogue selector (section \ref{section:Dialogue Selector}) to manage the conversation flow. Additionally, after entity recognition, the selected entity types are linked to specialized external knowledge sources.

\subsection{Entity Linking and Knowledge Base}
We utilize external public domain-specific databases to obtain additional information about the recognised entity. We currently utilise 4 different databases for various entity types:
\begin{itemize}
  \item TMDb\footnote{www.themoviedb.org} database - Movie and Person (in the movies topical context) types.
  \item LastFM\footnote{www.last.fm} database - Song and Person (in the music topical context) types.
  \item Goodreads\footnote{www.goodreads.com} database - Book and Person (in the books topical context) types.
  \item IGDB\footnote{www.igdb.com} database - Videogame entity type.
\end{itemize}
For these entity types, we query the corresponding database with the recognized entity and from the returned candidates, we then select the one with the highest relevance and popularity based on the database's search algorithm.  

Each linked entity is then stored in the short-term session context in the User Profile together with the reference to the corresponding knowledge database and the information received from the database with the initial query. In certain specific contexts, the entity is also stored in the long-term User Profile context that persists between sessions. This can be, for example, the user's favorite movie or the user's last read book. 

This allows the system to utilize the information gained in the initial request efficiently. Moreover, whenever there is a need to access the entity in the database again, using a different query to obtain more information is simplified by retaining the entity reference.

In general, we have found this approach more manageable and with significantly less overhead than creating and maintaining our own dedicated general-purpose knowledge base would lead to.

\section{Dialogue Management}\label{section:Dialogue Management}

The dialogue management in the latest iteration of the Alquist socialbot follows the basic principles outlined in the previous iterations of the system \cite{alquist2, alquist3}.
The conversation consists of several small scripted dialogues, where turn-by-turn interactions in the context of the flow are handled solely by the intent detection component of the system (described in detail in Section \ref{section:Intent}).

Each dialogue revolves around one primary prompt (an example prompt can be whether the user has a pet or what is the user's favorite part of a previously mentioned movie) and possibly a small number of follow-up interactions. 

The novel part of dialogue management within our system concerns primarily the selection of these dialogues. We have created a new component called Dialogue Selector, whose purpose is to select the most relevant dialogue following after the previous dialogue has concluded so that the context and the coherence of the conversation remain intact.
The system takes into account all previously mentioned information that is retained in the User Profile, allowing it to exhibit socially intelligent behavior.





\subsection{Dialogue Selector}
\label{section:Dialogue Selector}
For the dialogue selector to judge the relevance of the dialogues, each dialogue is described by three types of information.

\begin{itemize}
  \item A set of tags that represent what topics the dialogue touches on. For example, the dialogue \textit{Esport} is tagged as \textit{Games} and \textit{Sport}.
  \item A set of User Profile attributes that are relevant to the dialogue. For example, the dialogue \textit{Favorite Movie} has relevant attributes \textit{favoriteMovie, discussedMovie, likesMovies}.
  \item A necessary starting condition that determines whether it is possible to initiate the dialogue within the current context successfully. For example, the dialogue \textit{Favorite-Movie-Part} requires that there is a movie being discussed with the user, thus the attribute \textit{discussedMovie} cannot be empty. Additionally, a common necessary starting condition is that the dialogue has not yet been initiated during the current session.
\end{itemize}

The dialogue selector utilizes this information together with the current state of the user profile to select a relevant continuation in the following steps:
\begin{enumerate}
  \item If there is relevant trivia for the currently discussed entity/topic and no trivia has been mentioned for three or more dialogues; the system selects the corresponding trivia as the next dialogue. (Experiments in the previous years of Alexa Prize have shown including relevant trivia in the conversation improves the results of the bot \cite{alquist2}).
  \item From all dialogues included in the socialbot, the system filters out all dialogues whose necessary starting condition has not been fulfilled.
  \item If there are no remaining dialogues, the system selects the Neural Response Generator for the continuation of the conversation.
  \item If there are remaining dialogues, the system looks at the dialogue tags that have just finished and compares them to the tags of the available dialogues. It then discards all dialogues except the ones with the highest overlap.
  \item From the new pool of dialogues, the system looks at the relevant attributes of the dialogue that have just finished and compares them to the attributes of the available dialogues. It again discards all dialogues except the ones with the highest overlap.
  \item If the system finds a non-zero overlap either in the dialogue tags or dialogue attributes, it selects the new dialogue randomly from the final dialogue pool.
  \item If the overlap in both dialogue tags and attributes is zero, the system instead tries to find an overlap between the User Profile attributes relevant to this session and the attributes of the available dialogues. 
  \item If the system finds a non-zero overlap between the relevant User Profile attributes and the dialogue attributes, it selects the new dialogue randomly from the final dialogue pool. This usually means selecting a new conversation topic in which the user has previously shown interest.
  \item Finally, if no overlap is found, the system initiates a recommendation dialogue designed to help the user select a new topic of the conversation.
\end{enumerate}

\section{Trivia Selection} \label{section:Trivia Selection}
The experiments done in Alquist 2.0 \cite{alquist2} showed that trivia information (also called fun-facts) about the topics and entities is an essential part of the conversation. To find relevant trivia related to the discussed entity and the context of the conversation is a rather difficult task. In Alquist 2.0, the trivia selection procedure was implemented roughly as follows: The trivia facts were scraped from Reddit on a daily basis. During the runtime of the conversation, the trivia database was queried using only the value and type of the currently discussed entities. The most recent trivia was returned. 

This approach suffers from the selection of irrelevant trivia information in specific situations. We try to address this issue using a model estimating the similarity between the trivia text and the recent context of the conversation.

\subsection{Model}
The task of the model is formulated as an estimation of similarity between the text of the trivia and $ n $~most recent utterance--response pairs. We empirically estimated the value of $ n $ to 2. First, we need to create a dense representation of both the trivia text and the utterance--response pairs. Then we compute the cosine similarity of the vector representations and select the most similar trivia given the context of the dialogue turns. The full procedure can be described in the following steps.
\begin{enumerate}
    \item During the scraping phase, the trivia text is encoded, and the dense representation is stored.
    \item During runtime, a candidate trivia list is retrieved using a full-text search with the entity text as query.
    \item Context of $ n $~most recent utterance--response pairs is encoded.
    \item Cosine similarity between the context representation and each candidate is computed, and the most similar one is selected.
\end{enumerate}
We experimented with several model architectures for the encoders. We tried to use the encoders without any fine-tuning on task-specific data, and then we measured the improvement of those models if they were fine-tuned using relevant data. The model architectures we experimented with are shown in Table \ref{tab:trivia}.

\subsection{Datasets}
We manually annotated the turns from the conversations gathered during the previous versions of our bot and used the data for fine-tuning. Only the turns (and their respective contexts) which contained trivia were considered. We created a dataset with 350 turns where the trivia was mentioned. Each sample consists of one suitable piece of trivia and four negative examples (randomly selected trivia), a context (list of user utterance--bot response pairs), and an annotation of whether the trivia is relevant given the context or not. Three hundred samples were used for actual fine-tuning, and fifty samples were used for testing.

\subsection{Results and Discussion}
Table \ref{tab:trivia} shows the results using different encoder models. We experimented with both not fine-tuned and fine-tuned versions. Each model scored five candidate pieces of trivia. We also compared the results with the DialogRPT ranker used in the Neural Response Generator. The results are compared with a baseline approach---the final trivia is selected randomly.

\begin{table}[h!]
\centering
\caption{Trivia selection experiments results. The columns show how many times the most suitable trivia was scored as the most similar among the top 2 candidates, or among the top 3 candidates.}
\label{tab:trivia}
\begin{tabular}{llll} 
\hline
\textbf{Model}                & \textbf{Acc@1}         & \textbf{Acc@2}         & \textbf{Acc@3}          \\ 
\hline
Baseline             & 20\%          & 40\%         & 60\%          \\
\hline
Albert \cite{lan2019albert}  & 30\%          & 52\%         & 71\%           \\
Albert  (fine-tuned) & 34\%          & 54\%          & 74\%           \\
BERT \cite{devlin2019bert} (fine-tuned)   & 34\%          & 54\%          & 74\%           \\
Sentence-BERT \cite{reimers2019sentence}       & 49\%          & 71\%          & 85\%           \\
Sentence-RoBERTa \cite{reimers2019sentence}     & 61\%          & 83\%          & 92\%           \\
DPR \cite{karpukhin2020dense}                 & 20\%          & 42\%         & 72\%           \\
DPR  (fine-tuned)    & \textbf{65\%} & \textbf{84\%} & \textbf{93\%}  \\ 
\hline
DialogRPT \cite{gao2020dialogue}           & 14\%          & 31\%          & 53\%           \\
\hline
\end{tabular}
\end{table}

\section{Intent and Out-of-Domain Classification}
\label{section:Intent}

Following the concept of dialogue presented in Alquist 2.0 \cite{alquist2}, we design each dialogue as a tree structure. The tool described in \cite{alquist2,alquist3} is used for designing the dialogues. A crutial point in the conversation structure is where we expect the user input/user utterance. Each user utterance is then classified into a specific intent for which the dialogue designer manually writes training utterances. However, because of the complexity of language and the open-world assumption \cite{Keet2013_open_world_assumption}, the dialogue designer cannot incorporate each possible intent. Based on that, these user utterances for which the dialogue is not prepared are called out-of-domain. Therefore, the intent can be in-domain (ID) or out-of-domain (OOD). The ID intent is a user utterance for which the dialogue designers have prepared a response. Such a response is designed in a coherent and engaging conversational style. 

We have also incorporated the concept of hierarchy into our dialogue design and introduced two types of ID intents --- intents valid across all dialogues (global ID intent) and intents valid only in the specific context (local ID intent). However, despite the fact that the OOD detection has been receiving more attention lately \cite{marek2021ood_gan,tan2019ood_outofdomain}, the current datasets for evaluating the OOD performance \cite{larson-etal-2019-evaluation-clinc150,gangal2019likelihood_rostd,8683019_context_ood} are not designed for testing the hierarchical structure of our dialogues and contain mainly explicit commands. To solve this issue, we created our testing data from anonymized queries of real users. The following subsection describes our hierarchical model, performed experiments, and the obtained results.

\begin{figure}[h!]
    \centering
    \includegraphics[width=0.6\columnwidth]{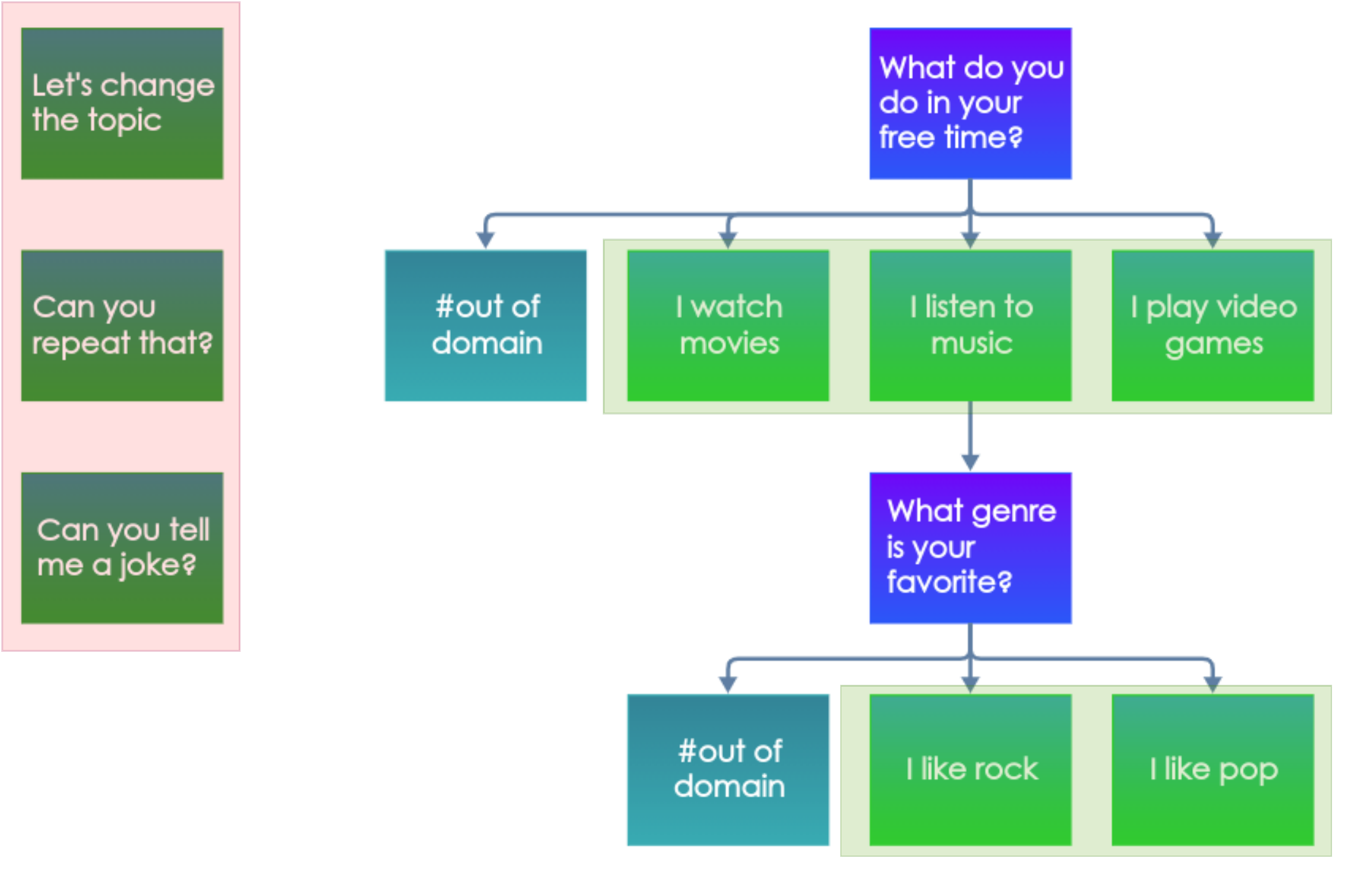}
    \caption{Hierarchical structure of dialogues. The dialogue consists of the nodes representing user inputs (green nodes) and bot responses (blue nodes). The chained nodes create a diverging flow of conversation. We omitted some of the nodes for clarity. 
    The user input nodes represent intents. There are two types of intents. The user utterance can be classified into one of the local intents (green boxes) specific to a certain turn of the dialogue, plus any global intent (red box) that can be applied in any turn of the dialogue. Moreover, the user utterance can be classified as out-of-domain. }
    \label{fig:stucture1}
\end{figure}

\subsection{Model}
\label{sub:model}

As described in the previous section, our intents create a hierarchical structure, which can be seen in \autoref{fig:stucture1}. The hierarchical structure provides the dialogue designer with modularity in creating the flow of the conversation but puts a significant emphasis on the effectiveness of the algorithms for intent classification. 

\begin{figure}[h!]
    \centering
    \includegraphics[width=0.8\columnwidth]{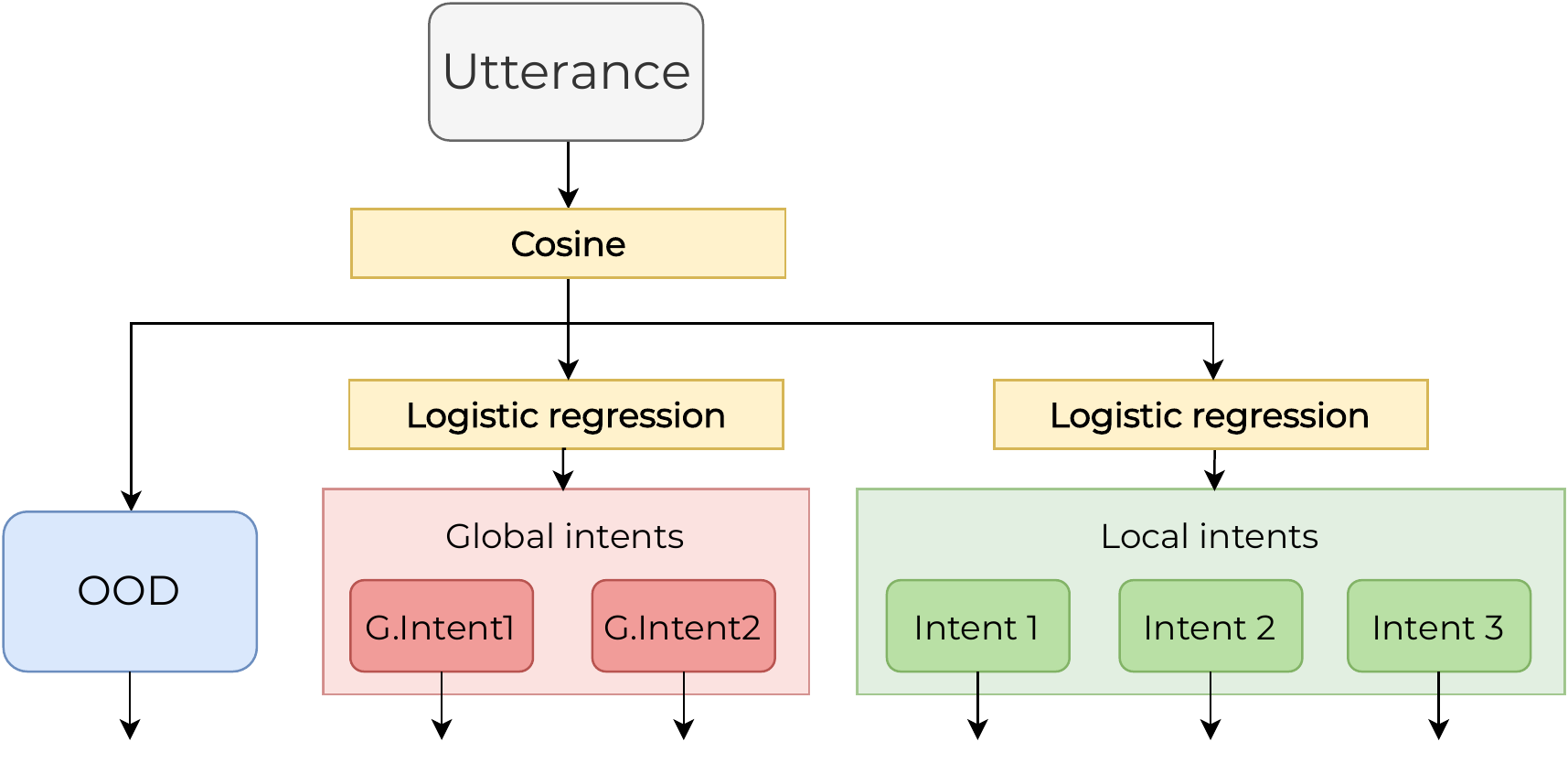}
    \caption{Classification algorithm for intent and OOD detection. The utterance is first classified by a cosine similarity into the class of local intents or the class of global intents. Next, the corresponding logistic regression makes the final intent classification. Additionally, cosine similarity can predict the OOD class if the similarity score falls bellow the threshold.}
    \label{fig:model}
\end{figure}

To allow the suggested modularity, we train a separate model for each level of the hierarchy. We show the whole classification system in \autoref{fig:model}. The system works in two steps. First, we need to determine which intent model in the hierarchy is appropriate --- local or global. We utilize cosine similarity over sentence embeddings between user query and train examples of each intent to classify utterances as local or global intents while prioritizing local intents. The priority is based on using a stable sorting algorithm over cosine similarities between sentence embedding of the user utterance and the examples of each intent. We also allow manual setting of the threshold for local and mainly for global intents leading to the filtering out of the intents if the cosine similarity is not high enough. In the second step, the utterance is classified into a specific intent by the corresponding logistic regression selected in the previous step. We use logistic regression because of the speed of its training and the proven performance in low-resource scenarios (see \autoref{fig:dependency}). Thus, the final intent classification is performed by logistic regression.

Additionally, the cosine similarity can filter out all ID classes (if the similarity score falls below the threshold). It will lead to the output of the OOD class. This is an approach similar to \cite{larson-etal-2019-evaluation-clinc150}. In addition, to make our system more robust, we include a dynamic threshold similar to \cite{shu2017doc_deep_open_classifier}. Our dynamic threshold is based on the arithmetic mean of similarities between the two closest train examples in each intent. This dual-threshold approach balances trade-offs between the manual control of the OOD sensibility and the robustness of the whole system. 

\vspace{0.5cm}
\subsection{Datasets}

We performed our analysis on a publicly available dataset as well as on manually labelled anonymized queries:

\begin{itemize}
    \item CLINC150 Dataset \cite{larson-etal-2019-evaluation-clinc150}
    \item ALQUIST 4.0 Dataset
\end{itemize}

A summary of the datasets is shown in \autoref{tab:alquist_data} and \autoref{tab:clinc150_data}. All samples in the ALQUIST 4.0 Dataset were carefully checked for consistency and drawn from aggregated and anonymized queries. The datasets generated during this study are not made available for privacy reasons. The CLINC150 Dataset was augmented to support the hierarchical structure --- the augmentation was performed as a random selection of 15 intents from the CLINC150 Dataset and then divided into two sets --- one set represents local intents, and the other represents global intents. The ratio was set to 1:3 to represent a typical situation in a dialogue. The split results in 4 local intents and 11 global intents --- a detailed overview is shown in \autoref{tab:clinc150_data}. The process was repeated 15 times. The results shown in the experimental section are average. 

\begin{table}[h!]
\centering
\caption{ALQUIST 4.0 Dataset}
\label{tab:alquist_data}
\begin{tabular}{c|c|c} 
\hline
\textbf{Type of utterances} & \textbf{Average Per Intent} & \textbf{Total Utterances}  \\ 
\hline
Local - Train               & 564                         & 3952                       \\ 
\hline
Global - Train              & 399                         & 2793                       \\ 
\hline
Local - Test                & 27                          & 193                        \\ 
\hline
Global - Test               & 49                          & 344                        \\ 
\hline
ODD - Test                  & 61                          & 429                        \\
\hline
\end{tabular}
\end{table}

\begin{table}[h!]
\centering
\caption{CLINC150 Augmented Dataset}
\label{tab:clinc150_data}
\begin{tabular}{c|c|c} 
\hline
\textbf{Type of utterances} & \textbf{Average Per Intent} & \textbf{Total Utterances}  \\ 
\hline
Local - Train               & 400                         & 6000                       \\ 
\hline
Global - Train              & 1100                        & 16500                      \\ 
\hline
Local - Test                & 120                         & 1800                       \\ 
\hline
Global - Test               & 330                         & 4950                       \\ 
\hline
ODD - Test                  & 1000                        & 1000                       \\
\hline
\end{tabular}
\end{table}

\vspace{0.8cm}
\subsection{Experiments}

Our analysis includes an inspection of the input features (embeddings). We include the following sentence embeddings:

\begin{itemize}
    \item Average of word embeddings FastText \cite{mikolov2018fasttext}
    \item Universal Sentence Encoder - Deep Average Network (USE-DAN) \cite{cer2018universal_sentence_encoder} 
    \item Universal Sentence Encoder - Transformed-encoder (USE-TRAN) \cite{cer2018universal_sentence_encoder}
\end{itemize}

The model described in \autoref{sub:model} is tested in two ways --- automatically and manually. The automatic evaluation is performed over our ALQUIST 4.0 Dataset (shown in \autoref{tab:alquist_data}) and the artificially hierarchical augmentation of the CLINC150 Dataset (shown in \autoref{tab:clinc150_data}). The manual evaluation was performed on aggregated data selected from anonymized user conversations with the socialbot. We collected all user utterances from these parts of dialogues and performed the human evaluation. The results can be seen in \autoref{tab:manual}. Besides evaluating the performance solely for the OOD detection, we looked at the performance of the local and global intents. The results are discussed in the following section.

\begin{table}[]
\centering
\caption{Manual evaluation on aggregated data}
\begin{adjustbox}{width=0.8\linewidth}
\begin{tabular}{ll|l|l|l} 
\hline
                 &                        & \multicolumn{1}{l}{True intent} & \multicolumn{1}{l}{}   &               \\
\textbf{}        &                        & \textbf{Local intent}           & \textbf{Global intent} & \textbf{OOD}  \\ 
\hline
Predicted intent & \textbf{Local intent}  & 3451                            & 23                     & 35            \\ 
\cline{2-5}
                 & \textbf{Global intent} & 0                               & 472                    & 0             \\ 
\cline{2-5}
                 & \textbf{OOD}           & 0                               & 15                     & 238           \\
\hline
\end{tabular}
\end{adjustbox}

\label{tab:manual}
\end{table}

\begin{figure}[h!]
    \centering
    \includegraphics[width=0.8\columnwidth]{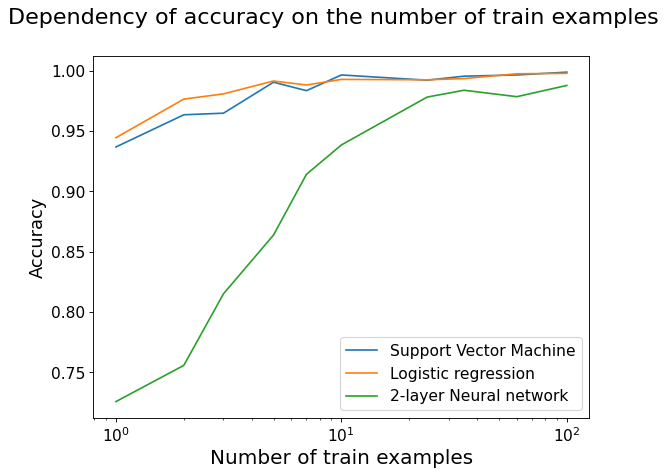}
    \caption{Dependency between the number of examples and the accuracy for the three most common classification models. Measurement was performed on the CLINC150 Dataset. The number of examples is in a logarithmic scale and was measured on a range from 1 to 100 train examples.}
    \label{fig:dependency}
\end{figure}

To select the most suitable model for the final intent classification, we measure the difference between the three most common classification models --- Logistic Regression, Support Vector Machine, and the 2-layer Neural Network. We focus on the necessary number of needed examples to achieve sufficient accuracy. The evaluation is shown in \autoref{fig:dependency} and highlights the problem of the neural network when dealing with low-resource scenarios.  The measurement was performed over CLINC150, randomly choosing five classes (our average number of intent classes for the intent model) and randomly choosing N examples. This procedure was repeated 25 times, and then shown values are average. We selected Logistic Regression as a model performing well in the low-resource scenario.

\subsection{Results and Discussion}

Taken together, our results (shown in \autoref{tab:result_alquist}) suggest a relationship between the type of embedding and the performance of the classification model. The difference in results can be explained by the word-order sensitivity of advanced embedding techniques. Another important aspect is the memory requirements and the speed of obtaining the embedding (shown in \autoref{tab:memory}). The results suggest the usage of USE-DAN as an appropriate embedding layer.

It should be mentioned that the performance on each level of the hierarchy remarkably differs between our two datasets. We believe that it is caused by the artificial augmentation of the CLINC150 Dataset and its unrealistic representation of the real-world use cases. We should also notice the higher performance for the local intent classification than for the global intent classification (notable mainly on the ALQUIST 4.0 Dataset). It is caused by the hierarchical structure of our model, which emphasizes the local intent over the global as was described in \ref{sub:model}. This is aligned with our experience that staying in the local context of the dialogue is beneficial for coherence. 

The performance of the OOD detection needs to be evaluated with respect to precision and recall. The high precision and lower recall indicate that the algorithm is suitable for classifying OOD in the conversational domain because we prefer false negative over false positive - it supports the consistency of the dialogue. In addition, we performed a human evaluation (shown in \autoref{tab:manual}) which demonstrates the performance on real-world data. It supports previously stated conclusions.

\begin{table}[h!]
\centering
\caption{Intent classification results}
\begin{adjustbox}{width=0.9\linewidth}
\begin{tabular}{ccccccccc} 
\hline
                                     &                      & \multicolumn{3}{c}{\textbf{CLINC150}}              & \textbf{} & \multicolumn{3}{c}{\textbf{ALQUIST 4.0 DATA}}       \\ 
\cline{3-5}\cline{7-9}
\textit{Embeddings}                  & \textit{Intent type} & \textit{Precision} & \textit{Recall} & \textit{F1} & \textit{} & \textit{Precision} & \textit{Recall} & \textit{F1}  \\ 
\hline
\multirow{3}{*}{\textbf{FastText}}   & Local                & 99.6\%             & 91.6\%          & 95.2\%      &           & 85.5\%             & 61.6\%          & 68.7\%       \\
                                     & Global               & 98.9\%             & 89.3\%          & 92.8\%      &           & 67.8\%             & 42.7\%          & 48.2\%       \\
                                     & OOD                  & 99.8\%             & 44.1\%          & 61.2\%      &           & 99.8\%             & 81.3\%          & 89.6\%       \\ 
\hline
\multirow{3}{*}{\textbf{USE - DAN}}  & Local                & 99.9\%             & 96.5\%          & 98.1\%      &           & 90.8\%             & 69.5\%          & 74.3\%       \\
                                     & Global               & 99.6\%             & 96.6\%          & 98.0\%      &           & 77.0\%             & 54.7\%          & 60.9\%       \\
                                     & OOD                  & 99.9\%             & 89.0\%          & 94.2\%      &           & 99.9\%             & 83.8\%          & 91.2\%       \\ 
\hline
\multirow{3}{*}{\textbf{USE - TRAN}} & Local                & 99.5\%             & 97.7\%          & 98.5\%      &           & 80.0\%             & 60.9\%          & 65.5\%       \\
                                     & Global               & 99.5\%             & 97.7\%          & 98.5\%      &           & 85.4\%             & 59.6\%          & 67.2\%       \\
                                     & OOD                  & 99.9\%             & 91.1\%          & 95.3\%      &           & 99.9\%             & 85.6\%          & 92.2\%       \\
\hline
\end{tabular}
\end{adjustbox}
\label{tab:result_alquist}
\end{table}

\begin{table}[h!]
\centering
\caption{Requirements of different embedding algorithms}
\begin{tabular}{l|ll} 
\hline
\textbf{Embeddings} & \textbf{Speed} & \textbf{RAM usage}  \\ 
\hline
\textbf{FastText}   & 200 it/s       & 190 MB              \\
\textbf{USE - DAN}  & 120 it/s       & 1765 MB             \\
\textbf{USE - TRAN} & 25 it/s        & 1650 MB             \\
\hline
\end{tabular}
\label{tab:memory}
\end{table}

\section{Neural Response Generator}\label{section: Neural Response Generator}
A Neural Response Generator (NRG) is a neural conversational response generation model trained on large conversational corpora. It generates a response based on the most recent turns of dialogue. We use such a model in Alquist in two settings. The neural response generator creates a response for out-of-domain user inputs, and it generates follow-up questions about trivia.

The motivation to use a Neural Response Generator for the out-of-domain (OOD) inputs is the following. We put the main content emphasis in Alquist 4.0 on hand-designed dialogues. Dialogues are represented as graphs (\autoref{fig:stucture1}). They consist of nodes representing the user inputs and the bot responses structured in diverging dialogue flows. Although they allow a human designer to create high-quality conversations with maximum control over the dialogue flow, the designer can not predict all possible flows the conversation can go through. We can detect the situations in which the user diverts from the predesigned flow by the out-of-domain detection. However, the problem of how to continue in the conversation in a meaningful way emerges. The fact that we can't predict all possible flows of the conversation also means that we can't design them. Neural response generators that can create a response on the fly based on the context of the dialogue and the user input are the way we solve the problem.


To enhance the conversation with interesting and surprising pieces of information, we use crawled trivia from Reddit\footnote{https://www.reddit.com/r/todayilearned/}. Trivia is a short sentence that contains interesting or surprising information. The trivia has a form of a statement. This property makes incorporating them into the dialogue in a conversational way challenging because statements don't encourage the user to continue in the conversation as questions do. The solution is to concatenate a follow-up question to the trivia. Because there is a large number of trivia, it is intractable to write a  follow-up question for each piece. Thus, we use a Neural Response Generator to create follow-up questions.

The practical application of a Neural Response Generator faces several challenges. First, the generated responses have to be quality enough. Second, the response must be generated quickly enough to be applicable to a conversation in real-time and with reasonable computational resources. Additionally, the lack of control over the generated response is a factor that limits the application of the Neural Response Generator in combination with hand-designed dialogue content. One of the most significant obstacles we identified is randomness, in which the generator produces questions and statements. The main problem is that questions encourage the user to respond, whereas statements do not. Put simply, if the generator produces a question, the system should not continue in the dialogue immediately. Instead, the user should have space to respond to it. On the other hand, statements can serve as a connection between generated and hand-designed content. We demonstrate the hypothetical dialogue situations in \autoref{tab:NRG-statement-question}, in which the user has an opportunity to answer a generated question, and a statement is used as a reaction and connection to hand-designed dialogue. We wanted to use a Neural Response Generator in both mentioned situations. Thus, we focused on developing the control mechanism of the model that specifies whether we want to generate a question or a statement. Therefore, all the factors ranging from the quality of responses to the inference time and the ability to control output are the criteria that influenced the evaluations we performed and the model we selected.

\begin{table}[]
\centering
\caption{A hypothetical dialogue situation, in which a statement--question control mechanism of the NRG is used. The first user input is identified as out-of-domain. The Neural Response Generator produces a question that is followed by the user utterance. That is followed by a generated statement that transits into a hand-designed dialogue.}
\begin{tabular}{r|l}
\hline
Alquist (Hand-designed dialogue): & \textit{What do you do in your free time?}    \\  
\hline User (Out-of-domain):    & \textit{I draw pictures.}                      \\ \hline
Alquist (NRG-Question):    & \textit{What kind of pictures do you like to draw the most?}                     \\ \hline
User:                      & \textit{I usually draw portraits}      \\ \hline
Alquist (NRG-Statement):    & \textit{I see. Portraits are hard to make.}  \\ \hline
Alquist (Hand-designed dialogue): & \textit{Do you visit galleries?}    \\    \hline         
\end{tabular}
\label{tab:NRG-statement-question}
\end{table}

\subsection{Model}
We selected DialoGPT \cite{zhang2020dialogpt} for our experiments. It is a large, tunable neural conversational response generation model trained on 147M conversation-like exchanges extracted from Reddit comment chains over a period spanning from 2005 to 2017. DialoGPT is based on GPT-2 \cite{radford2019language}. The GPT-2 model is built out of the transformer language model \cite{vaswani2017attention} and leverages a stack of masked multi-head self-attention layers. The text generated either from scratch or based on a user-specific prompt is realistic-looking. DialoGPT attains a performance close to human both in automatic and human evaluation in single-turn dialogue settings. There are three sizes of the model: \emph{small} with 117M parameters, \emph{medium} with 345M parameters, and \emph{large} with 762M parameters. 

DialoGPT optimizes the conditional probability 
\[p(T|S) = \prod_{n=m+1}^{N} p(x_n|x_1, ... , x_{n-1})\]
where we concatenate all dialogue turns within a dialogue context into a long text \(x_1,...,x_N\) (\(N\) is a sequence length). Each dialogue turn is followed by a special \emph{end-of-speech} token. We denote the dialogue history as \(S = x_1, ..., x_m\) and the target sentence as \(T=x_{m+1},...,x_N\). 

To have control over whether the model generates a question or a  statement, we modified the original DialoGPT. We introduced special tokens to the beginning of the input to DialoGPT. Special tokens \emph{QUESTION} or \emph{STATEMENT} are prepended to the dialogue context. They give information to the model, whether our desired response should be a statement or a question. Thus, we optimize the conditional probability
\[p(T|S) = \prod_{n=m+1}^{N} p(x_n|\textup{\textit{STATEMENT}},x_1, ... , x_{n-1})\]
if \(T\) is a statement, and 
\[p(T|S) = \prod_{n=m+1}^{N} p(x_n|\textup{\textit{QUESTION}},x_1, ... , x_{n-1})\]
if \(T\) is a question.

DialoGPT can produce several candidate responses. And because some replies are more engaging than others, spawning more follow-up interactions, we have to select the optimal one. We employed DialoRPT \cite{gao2020dialogue} for this task. DialoRPT is a set of GPT-2 based ranking models trained on 133M pairs of human social media feedback data (number of replies and upvotes) built for feedback prediction. There are five types of rankers: \emph{updown} that predicts how likely the response gets the most upvotes, \emph{width} that predicts how likely the response gets the most direct replies, \emph{depth} that predicts how likely the response gets the longest follow-up thread, \emph{human\_vs\_rand} that predicts how relevant the response is for the given context, and \emph{human\_vs\_machine} that predicts how likely the response is human-written rather than machine-generated. DialoRPT takes the generated candidate responses and produces a score for each. We select the response that has the largest score. This way, we get the best response according to the selected DialoRPT model.

\subsection{Datasets}
Because the quality of data influences the quality of responses, we used the following datasets for experiments:
\begin{itemize}
    \item \textbf{Alquist Dialogue Graphs}
    
    Alquist Dialogue Graphs consist of the dialogue graphs introduced in \cite{alquist2} and represents the human-designed dialogue flows of the socialbot. It consists of 640k dialogues generated out of 80 dialogue graphs. Thus, those dialogues have a relatively small semantic diversity. 
    
    \item \textbf{Topical-Chat \cite{Gopalakrishnan2019}}
    
    Topical-Chat is a knowledge-grounded human-human conversation dataset where the underlying knowledge spans eight broad topics, and conversation partners don’t have explicitly defined roles. It consists of 10k conversations and 235k utterances.
    
    \item \textbf{EmpatheticDialogues \cite{rashkin2019towards}}
    
    EmpatheticDialogues is a dataset of 25k dialogues grounded in situations prompted by specific emotion labels.
    
    \item \textbf{DailyDialog \cite{li2017dailydialog}}
    
    DailyDialog is a multi-turn dialogue dataset that reflects daily communications and covers various topics about everyday life. The dataset is manually labelled with communication intention and emotion information. It contains 13k dialogues.
    
    \item \textbf{Merged datasets}
    
    Dataset consisting of all dialogues taken from Alquist Dialogue Graphs, Topical-Chat, DailyDialog, and Persona-Chat \cite{zhang2018personalizing}. It represents the most extensive mix of various conversational styles we used.
\end{itemize}

Some of the used datasets contain additional annotations, like emotions, knowledge, situation description, or information about the speaker. However, we decided to not use the annotations in our experiments because those annotations are not available in our system in real traffic.

On the other hand, we modified the dataset to include special \emph{QUESTION} and \emph{STATEMENT} tokens. First, we split the turns into sentences by the NLTK \cite{bird2006nltk} sentence tokenizer. Second, Standford CoreNLP \cite{manning2014stanford} annotated each sentence as a statement or a question. Third, if the turn contained both statements and questions, we divided the turn into several turns where each turn contains sentences of the same type, and we did not modify the order of sentences. For example, we split the turn \(S_1S_2Q_3Q_4\) (where \(S_i\) denotes a statement sentence consisting of tokens \(x_1,...x_s\) and \(Q_j\) denotes a question sentence consisting of tokens \(x_1,...x_q\)) into two turns: \(S_1S_2\) and \(Q_3Q_4\). Lastly, we label all turns consisting of statements by a special token \emph{STATEMENT} and all turns consisting of questions by a special token \emph{QUESTION}. Using these steps, we create a dataset in which each turn consists of either questions or statements and is labelled by a special token utilized during training. 

\begin{figure}[h!]
    \centering
    \includegraphics[width=\columnwidth]{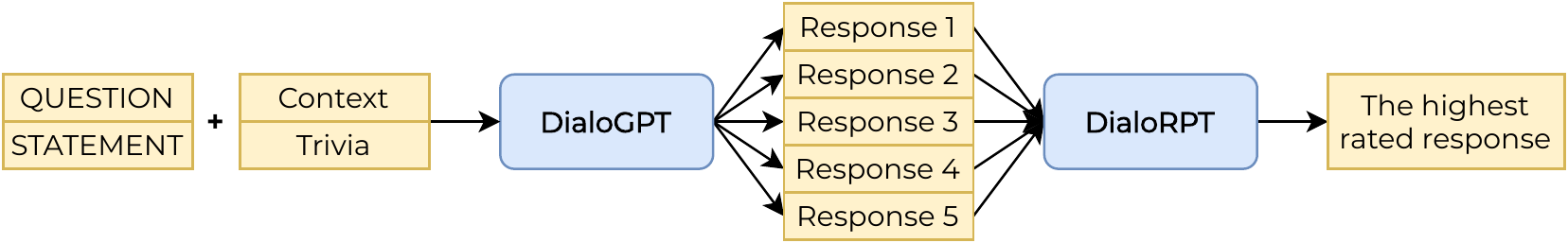}
    \caption{The diagram of the neural response generator. The input concatenation of the \emph{QUESTION} or \emph{STATEMENT} special token and the string of dialogue context or trivia. DialoGPT processes the input and generates several response variants. DialoRPT rates responses, and the highest-rated response is returned as a result.}
    \label{fig:stucture}
\end{figure}

\subsection{Evaluations}
We evaluated \emph{small}, \emph{medium}, and \emph{large} DialoGPT models (with the \emph{STATEMENT} and \emph{QUESTION} token modification) trained on five datasets. The models used beam search with five beams. Each of the models generated five alternative responses that were ranked by the \emph{updown} DialoRPT model and we took the top response for evaluation.

We evaluated the models in three metrics. Question/Statement accuracy evaluated models ability to generated questions and statements correctly based on the specification of the former or the latter. We fixed 1000 three turn long contexts as inputs. For each of them, we generate five responses with the \emph{QUESTION} token and five with the \emph{STATEMENT} token. Next, Stanford CoreNLP annotated all generated responses as a question or a statement. In total, we have 10,000 examples for which we compute the accuracy.

Next, we performed the human evaluation of the model outputs. We performed an evaluation for OOD and trivia. For OOD, we took 100 dialogue contexts (3-5 utterances long) and generated the following response in the form of a question. The responses were manually labelled as \emph{OK} or not. Annotation \emph{OK} labels the response that correctly progresses in the conversation flow. Responses that we did not label as \emph{OK} were \emph{unrelated} to the given context, factually \emph{untrue} (\textit{"Do you know that whales are the largest birds?"}), or the information in them was \emph{repeated} in the given context (a user inputs \textit{"I have never been abroad."} and the generated response is \textit{"Have you ever been to Canada?"} for example). Those are the three most significant issues we identified in the responses generated by NRG. 

For trivia, we took 100 pieces of trivia crawled from Reddit. We used the trivia as an input to the DialoGPT model, and we generated the follow-up question. We also labelled responses as \emph{OK} or not. We further divided Not \emph{OK} responses into \emph{unrelated} to the given trivia or factually \emph{untrue}. We did not use the \emph{repeated} class as the only input to the model was the trivia. Thus, the model could not know that the given question was asked in the previous turns of the conversation.

We evaluated the inference time of the models as well, as it is a crucial parameter for the models to be applicable to conversations in real-time. We evaluated the models on CPU as well as GPU for several numbers of generated responses. The CPU was Intel(R) Xeon(R) Platinum 8175M CPU~@~2.50GHz used in AWS EC2 instance m5.2xlarge. The GPU was Tesla T4 used in AWS EC2 instance g4dn.4xlarge. We fixed the input to the model and averaged the time of 20 inferences.

\subsection{Results}
\autoref{tab:NRG_quality} presents the comparison of the \emph{small}, \emph{medium}, and \emph{large} DialoGPT models trained on five dialogue datasets \cite{Hylova2021report}. We can notice that all models on all datasets possess a good ability to produce questions or statements based on the desired output. The only dataset on which models do not get over a 90\% threshold is EmphateticDialogues. 

Next, we can notice that the larger the model is, the better responses it produces based on human evaluation on both OOD and trivia. The large DialoGPT model trained on EmpatheticDialogues produces the best responses for out-of-domain. The large DialoGPT model trained on Topical-Chat produces the best follow-up questions for trivia.

We also present the results of the error analysis in \autoref{tab:NRG_errors} \cite{Hylova2021report}, where we assigned each generated response to one of the three error classes (\emph{unrelated}, \emph{untrue}, and \emph{repeated}) we identified in the case of OOD and two error classes (\emph{unrelated}, and \emph{repeated}) in the case of funfacts.

We can see that the two biggest issues for all models in the case of OOD are \emph{unrelated} and \emph{repeated} responses. The situation is similar for trivia. The bigger problem than \emph{untrue} are \emph{unrelated} responses. To the best of our knowledge, there is no easy way to solve \emph{unrelated} responses. The \emph{repeated} responses might be filtered using semantic text similarity, but further research is needed. The \emph{untrue} responses are mainly a problem of the Topical-Chat dataset because of its nature. It is a dataset where two dialogue agents have conversations about trivia. Thus, we hypothesise that the model learns to generate responses that resemble trivia, but the model is not powerful enough to learn true trivia only. However, we identified that the model tends to start such responses by a phrase \emph{Did you know}. Thus we can use string matching to remove such responses in practice.

\autoref{tab:nrg-inference} presents the results of the experiment evaluating the inference times of \emph{small}, \emph{medium}, and \emph{large} models \cite{Hylova2021report}. The two obvious facts we can notice are that the larger the model is and the more variants it generates, the longer the inference time is. We can also see that the inference time is significantly shorter on GPU than on CPU. We face a dilemma here. On the one hand, we want as good responses as possible, we also want several of them to utilize the DialoRPT model, but we want to produce them in the shortest time possible too. We selected 400 ms as a time threshold we do not want to cross as more time poses a noticeable time delay in the conversation when we sum the processing time of the rest of the system's components. Next, we decided that we want as good responses as possible. Because the GPU cost was manageable for us, we selected the largest model running on GPU, generating five variants of response. Later in the competition, we switched to three variants because the processing time of the rest of the system increased due to larger complexity. Further research is needed to make inference times shorter or make the model tractable to run on the CPU (machines with CPUs only have cheaper hourly rate than machines with GPUs on cloud services usually).

\begin{table}[h!]
\centering
\caption{Results of NRG models}
\begin{tabular}{llrrrr}
\hline
\multirow{2}{*}{\textbf{Model}} & \multirow{2}{*}{\textbf{Dataset}} & \multicolumn{1}{l}{} & \multicolumn{1}{l}{} & \multicolumn{2}{c}{\textbf{OK}}   \\ \cline{5-6} 
                                &                                   & \textbf{Perplexity}  & \textbf{Q/S Acc.}    & OOD  & Funfacts \\ \hline
                                & Alquist Dialogue Graphs           & 11.89                & 96.1\%              & 50\%          & 62\%              \\
                                & DailyDialog                       & 14.72                & 98.5\%              & 53\%          & 38\%              \\
Small                           & Merged datasets                   & 22.90                & 95.9\%              & 47\%          & 73\%              \\
                                & Topical-Chat                      & 27.11                & 96.5\%              & 36\%          & 69\%              \\
                                & EmpatheticDialogues               & 20.45                & 95.7\%              & 53\%          & 49\%              \\ \hline
                                & Alquist Dialogue Graphs           & \textbf{6.88}        & 91.8\%              & 58\%          & 41\%              \\
                                & DailyDialog                       & 12.12                & 95.4\%              & 64\%          & 62\%              \\
Medium                          & Merged datasets                   & 18.32                & 95.9\%              & 57\%          & 82\%              \\
                                & Topical-Chat                      & 20.84                & 98.4\%              & 37\%          & 80\%              \\
                                & EmpatheticDialogues               & 16.51                & 76.2\%              & 73\%          & 69\%              \\ \hline
                                & Alquist Dialogue Graphs           & 7.13                 & 92.1\%              & 65\%          & 49\%              \\
                                & DailyDialog                       & 11.69                & 96.6\%              & 62\%          & 62\%              \\
Large                           & Merged datasets                   & 17.45                & 98.4\%              & 71\%          & 81\%              \\
                                & Topical-Chat                      & 20.46                & \textbf{99.0\%}     & 43\%          & \textbf{93\%}     \\
                                & EmpatheticDialogues               & 16.56                & 84.5\%              & \textbf{77\%} & 72\%              \\ \hline
\end{tabular}%

\label{tab:NRG_quality}
\end{table}

\begin{table}[h!]
\centering
\caption{Error analysis of NRG models}
\resizebox{\textwidth}{!}{%
\begin{tabular}{llrrrrrrrr}
\hline
\multirow{2}{*}{\textbf{Model}} & \multirow{2}{*}{\textbf{Dataset}} & \multicolumn{4}{c}{\textbf{OOD}}                            &  & \multicolumn{3}{c}{\textbf{Funfacts}}       \\ \cline{3-6} \cline{8-10} 
                                &                                   & OK            & Unrelated  & Untrue    & Repeated      &  & OK            & Unrelated  & Untrue    \\ \hline
                                & Alquist Dialogue Graphs           & 50\%          & 26\%         & 2\%          & 22\%          &  & 62\%          & 34\%         & 4\%          \\
                                & DailyDialog                       & 53\%          & 13\%         & 2\%          & 32\%          &  & 38\%          & 61\%         & 1\%          \\
Small                           & Merged datasets                   & 47\%          & 22\%         & 2\%          & 29\%          &  & 73\%          & 21\%         & 6\%          \\
                                & Topical-Chat                      & 36\%          & 25\%         & 14\%         & 25\%          &  & 69\%          & 9\%          & 22\%         \\
                                & EmpatheticDialogues               & 53\%          & 23\%         & \textbf{0\%} & 24\%          &  & 49\%          & 51\%         & \textbf{0\%} \\ \hline
                                & Alquist Dialogue Graphs           & 58\%          & 21\%         & 5\%          & 16\%          &  & 41\%          & 52\%         & 7\%          \\
                                & DailyDialog                       & 64\%          & 11\%         & \textbf{0\%} & 25\%          &  & 62\%          & 38\%         & \textbf{0\%} \\
Medium                          & Merged datasets                   & 57\%          & 19\%         & 3\%          & 21\%          &  & 82\%          & 11\%         & 7\%          \\
                                & Topical-Chat                      & 37\%          & 10\%         & 34\%         & 19\%          &  & 80\%          & \textbf{4\%} & 16\%         \\
                                & EmpatheticDialogues               & 73\%          & \textbf{9\%} & 3\%          & 15\%          &  & 69\%          & 31\%         & \textbf{0\%} \\ \hline
                                & Alquist Dialogue Graphs           & 65\%          & 18\%         & 5\%          & 12\%          &  & 49\%          & 47\%         & 4\%          \\
                                & DailyDialog                       & 62\%          & 17\%         & 3\%          & 18\%          &  & 68\%          & 32\%         & \textbf{0\%} \\
Large                           & Merged datasets                   & 71\%          & 12\%         & 2\%          & 15\%          &  & 81\%          & 14\%         & 5\%          \\
                                & Topical-Chat                      & 43\%          & 21\%         & 21\%         & 15\%          &  & \textbf{93\%} & 5\%          & 2\%          \\
                                & EmpatheticDialogues               & \textbf{77\%} & 11\%         & 2\%          & \textbf{10\%} &  & 72\%          & 21\%         & 7\%          \\ \hline
\end{tabular}%
}
\label{tab:NRG_errors}
\end{table}

\begin{table}[h!]
\centering
\caption{Inference times of NRG models}
\begin{tabular}{llrrrrrr}
\hline
\multirow{2}{*}{\textbf{Device}} & \multirow{2}{*}{\textbf{Model}} & \multicolumn{6}{c}{\textbf{Number of generated responses}}                           \\ \cline{3-8} 
\textbf{}       & \textbf{}      & 1 & 2     & 3       & 5       & 10      & 20 \\ \hline
                & Small          & 247 ms     & 307 ms & 371 ms   & 435 ms   & 549 ms   & 931 ms                  \\
CPU             & Medium         & 438 ms     & 525 ms & 745 ms   & 854 ms   & 1,233 ms & 2,219 ms                \\
                & Large          & 673 ms     & 980 ms & 1,177 ms & 1,488 ms & 2,102 ms & 4,057 ms                \\ \hline
                & Small          & 109 ms     & 119 ms & 126 ms   & 133 ms   & 179 ms   & 273 ms                  \\
GPU             & Medium         & 164 ms     & 174 ms & 208 ms   & 237 ms   & 304 ms   & 432 ms                  \\
                & Large          & 186 ms     & 271 ms & 305 ms   & 347 ms   & 465 ms   & 726 ms                  \\ \hline
\end{tabular}%

\label{tab:nrg-inference}
\end{table}

\subsection{Examples}
We present the selected generated responses by the best performing models in this section. It is a large DialoGPT trained on EmpatheticDialogues for OOD and large DialoGPT trained on Topical-Chat for trivia. 

\autoref{tab:ood-examples} presents generated questions for OOD. We kept only the last OOD input from the three-turn context that the model receives as input for clarity in \autoref{tab:ood-examples}. We can notice that the model generates relevant responses in the selected examples that contain the mentioned entities (cereal, video games, or anime). However, part of the generated questions are statements. This fact was hinted in the Question/Statement accuracy on the EmpatheticDialogues dataset in \autoref{tab:NRG_quality}. Although it is an error, we do not consider it as critical because the generated response continues the dialogue appropriately.

\autoref{tab:funfact-examples} presents the generated questions about trivia. We can notice that in the first example the model understands that the words \emph{book} and \emph{read} are related, and for the book-related trivia generates a question \emph{Have you read it?} that correctly addresses the user too. We can observe a similar phenomenon in the next three examples, where the model additionally understands that \emph{Brazil}, \emph{Bill Murray}, and \emph{Caesar salad} are the primary entities of trivia. Finally, in the second to last example, we can see that the model also addresses itself by a philosophically-looking question. The generated follow-up questions of this model look reasonable, and we hypothesise that it is because of the constrained domain of trivia.

\begin{table}[h!]
\caption{Examples of generated responses for OOD. We kept only the last OOD input from the three-turn context for clarity.}
\begin{tabularx}{\textwidth}{X|X} \hline
\textbf{OOD Input}                                                                                                                                                                                                                                                          & \textbf{Generated Response}            \\ \hline
I had cereal for breakfast                                                                                                                    & What kind of cereal did you have?             \\ \hline
Why are you talking to me so much this morning                                                                   & Why do you think I'm talking to you so? \\ \hline
I wanna take my mind off and talk about video games                                                                                                                           & Do you have a favorite video game?      \\ \hline
anime songs & What kind of anime are you into?     \\ \hline
are you a mystic                                                                & No, I am not. I do not believe in magic.               \\ \hline
what would happen to a computer if you pulled the plug                                                                                                                                  & I don't know, I don't think it would happen. I think the computer would die.     \\   \hline
\end{tabularx}%
\label{tab:ood-examples}
\end{table}

\begin{table}[h!]
\caption{Examples of generated follow-up questions for trivia.}
\begin{tabularx}{\textwidth}{X|l} \hline
\textbf{Trivia}                                                                                                                                                                                                                                                          & \textbf{Generated Question}            \\ \hline
The first book ever written using a typewriter was The Adventures of Tom Sawyer. Mark Twain must have had a lot of ink and a lot of patience.                                                                                                                    & Have you read it?             \\ \hline
In Brazil, school starts at 7 a.m. and is over by noon so children can go home and have lunch with their parents. Having meals together as a family is an important part of Brazilian culture.                                                                   & Have you ever been to Brazil? \\ \hline
Bill Murray doesn't have an agent. You can leave him a voicemail message, hoping that he'll like the project and will get back to you.                                                                                                                           & Do you like Bill Murray?      \\ \hline
Caesar salad originated from a Mexican city. Restaurateur Chef Caesar Cardini was hosting a 4th of July celebration. He ran out of food, and simply threw the ingredients he had lying around the kitchen into a unique salad. Obviously, it was a huge success. & Do you like Caesar salad?     \\ \hline
In 2016, Mozart sold more CDs than Beyoncé. In fact, Mozart sold the most CDs in 2016, beating out Adele, Drake, and Beyoncé, even though all of those artists had Grammy-winning hits that year.                                                                & I wonder why?                 \\ \hline
Dogs’ sense of smell is about 100, 000 times stronger than humans’. However, they have only one-sixth our number of taste buds.                                                                                                                                  & Are you a dog person?     \\   \hline
\end{tabularx}%
\label{tab:funfact-examples}
\end{table}

\subsection{NRG Conclusion}
We applied a Neural Response Generator to Alquist to generate responses for user inputs that are out-of-domain of our hand-designed dialogues and to generate follow-up questions to trivia to encourage a user to continue in the dialogue. We selected the DialoGPT model that generates several response variants, followed by the DialoRPT model that ranks them.  A lack of control over the model's output, mainly whether the model generates a statement or a question, made it challenging to incorporate the model into handmade dialogues. For this reason, we modified DialoGPT to condition the response on special tokens \emph{STATEMENT} and \emph{QUESTION}. 

The quality of responses, ability to control the output, and inference time were crucial in our system. Thus, we performed experiments to evaluate those properties. We used Alquist Dialogue Graphs, DialyDialog, Topical-Chat, EmpatheticDialogues, and Merged datasets for evaluations. We modified the datasets by dividing their turns and labelling them by the \emph{STATEMENT} and \emph{QUESTION} labels. We evaluated \emph{small}, \emph{medium}, and \emph{large} DialoGPT models. The experiments showed that the best performance on the OOD task achieved a large DialoGPT model trained on the EmpatheticDialogues dataset and the best performance on trivia achieved a large DialoGPT model trained on Topical-Chat dataset. The experiments also demonstrated that the question/statement accuracy was above 90\% for most models which we consider a good result.

From the point of view of inference time, the best option still tractable is the large DialoGPT model that generates five variants of responses below 400 ms using GPU. Further research is needed to shorten this time and to make the model tractable on the CPU.

The examples of generated responses show that the model can generate relevant responses that contain the entities mentioned in the input to the model. In the case of follow-up trivia questions, it understands the entity (\emph{book}) and the related activity (\emph{to read}), can utilize the primary entity of trivia and correctly asks for user's preferences.

\section{Conclusion}
In this paper, we have presented a new iteration of the Alquist socialbot system competing in the Alexa Prize Grand Challenge 4. 

In order to actively engage the user in the conversation, we have developed Skimmer, a component that learns about the user from their messages and fills in information in their User Profile. We are then able to utilize what we have learned about the user's interests and personality further in the conversation, making it evident that the socialbot is invested in learning more about the user, remembers their preferences, and takes them into account during the conversation.

We have introduced the out-of-domain query detection as a core functionality of our system. This allows us to hand over more control of the conversation to the user, which makes the socialbot seem more responsive to the user queries. However, it also introduces a problem of how to handle the OOD responses. 

In order to keep the conversation engaging even when moving in the  unexpected direction, we have also integrated two generative models into the system based on DialoGPT. We utilize neural response generation in two different contexts within the conversation. Firstly, the Neural Response Generator triggers when the user input is classified as OOD. Secondly, we utilize the NRG to generate a follow-up prompt when presenting the user with trivia relevant to the currently discussed topics.

We have shown that in both cases the NRG is able to generate a relevant continuation of the conversation. This allows the system to efficiently handle unexpected responses and previously unseen information and integrate them into the conversation while maintaining the flow of the conversation and its coherence.

\medskip

\small
\printbibliography

\end{document}